%% file: iclr2025_conference.tex
\documentclass{article} %
\usepackage{iclr2025_conference,times}

\input{math_commands.tex}

\usepackage{hyperref}
\usepackage{url}
\usepackage{arydshln}
\usepackage{graphicx}
\usepackage{pifont}
\usepackage{bbm}

\iclrfinalcopy

\title{MonoFormer: One Transformer for Both
Diffusion and Autoregression}

\author{ \textbf{Chuyang Zhao$^{1\dag}$
  \quad Yuxing Song$^{1\dag}$
  \quad Wenhao Wang$^{2}$
  \quad Haocheng Feng$^{1}$
  \quad Errui Ding$^{1}$} \vspace{2mm} \\
  \textbf{
  \hspace{30mm}Yifan Sun$^{1*}$
  \quad Xinyan Xiao$^{1*}$
  \quad Jingdong Wang$^{1}$\thanks{Corresponding authors, $^{\dag}$Equal contribution}
  \vspace{4mm}
  }\\
  \hspace{30mm}$^1$Baidu VIS~~\quad\quad $^2$University of Technology Sydney
}

\newcommand{\method}{MonoFormer}

\newlength\savewidth\newcommand\shline{\noalign{\global\savewidth\arrayrulewidth
\global\arrayrulewidth 1pt}\hline\noalign{\global\arrayrulewidth\savewidth}}

\begin{document}

\maketitle

\begin{abstract}
Most existing multimodality methods
use
separate backbones for autoregression-based discrete text generation and diffusion-based 
continuous visual generation,
or the same backbone 
by discretizing the visual data
to use autoregression
for both text
and visual generation.
In this paper,
we propose to study
a simple %
idea:
share one transformer
for both autoregression and diffusion.
The feasibility comes from
two main aspects:
(i) Transformer is successfully
applied to diffusion for visual generation,
and (ii) transformer training
for autoregression and diffusion 
is very similar,
and the difference merely lies in that
diffusion uses bidirectional attention mask
and autoregression uses
causal attention mask.
Experimental results 
show that our approach
achieves
comparable image generation performance to current state-of-the-art methods as well as maintains
the text generation capability.
The project is publicly available at \url{https://monoformer.github.io/}.

\end{abstract}

\section{Introduction}

Diffusion models are popular 
for image generation
and other continuous data.
It is a probabilistic approach to modeling continuous data,
which creates samples by simulating the diffusion process, gradually adding and removing noise from data.
Diffusion is initially studied in the pixel space
for visual generation~\citep{ho2020denoising}.
Latent diffusion models~\citep{rombach2022high} performs
the diffusion in the latent representation space,
and are now commonly used
in many well-known models,
such as Stable Diffusion~\citep{rombach2022high} and DiT~\citep{peebles2023scalable}.

Autoregressive models
are dominant in large language models.
The basic idea is to predict the discrete tokens
one by one.
It is also widely studied for visual generation
by discretizing the image patches
through VQ-VAE~\citep{van2017neural} or dVAE~\citep{Ramesh21}.
Autoregression is advantageous
in building a unified transformer
for multi-modality understanding and generation
models~\citep{sun2024autoregressive,liu2024visual,team2024chameleon}.
Unfortunately,
it does not benefit from
recent advances in diffusion.
Autoregression and diffusion
are studied
in parallel
and often learned in different models.
Some attempts for combining them
for
multi-modality understanding and generation
models
adopt two separate networks
for text generation and visual generation, respectively~\citep{sun2023emu,lian2023llm}.

\begin{figure}[t]
  \centering
  \footnotesize
  \includegraphics[scale=0.95]{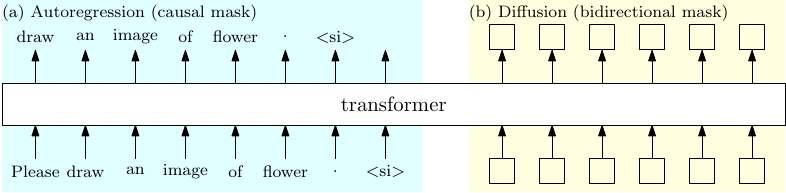}
   \caption{Our approach 
   MonoFormer
   trains the autoregressive transformer
   and the diffusion transformer,
   which share the weights,
   and uses causal attention mask
   and bidirectional attention mask, respectively.
   During training,
   the input of the transformer
   for autoregression
   is the text token embeddings,
   and the output is embeddings
   that are further processed for text generation.
   The input for diffusion
   is the noised latent embeddings,
   and the output is embeddings
   that are used to predict the noise.}
   \label{fig:OT}
\end{figure}

In this paper,
we aim at building 
and training one transformer
for both autoregression and diffusion.
The proposed approach
is named as MonoFormer
and is illustrated in Figure~\ref{fig:OT}.
The idea is very simple
and inspired by 
the success
of using transformer for diffusion~\citep{peebles2023scalable}
for image generation,
as well as 
the below-discussed observations
about the transformer for autoregression and
diffusion.
The main difference in
training the transformer
is that
autoregressive transformer
adopts a causal attention mask 
and diffusion transformer
does not mask any position,
or uses a bidirectional attention mask.
On the other hand,
the transformer receives continuous embeddings,
e.g., text token embedding
or image encoding,
and outputs continuous embeddings
for subsequent text token prediction
and image decoding.
Thus,
it is feasible to 
learn one transformer
for both discrete autoregression
and continuous diffusion.
We demonstrate
the idea
by training a single transformer
that is shared by
autoregression
for text generation
and diffusion 
for image generation.

\begin{figure}
    \centering
    \includegraphics[width=1.0\linewidth]{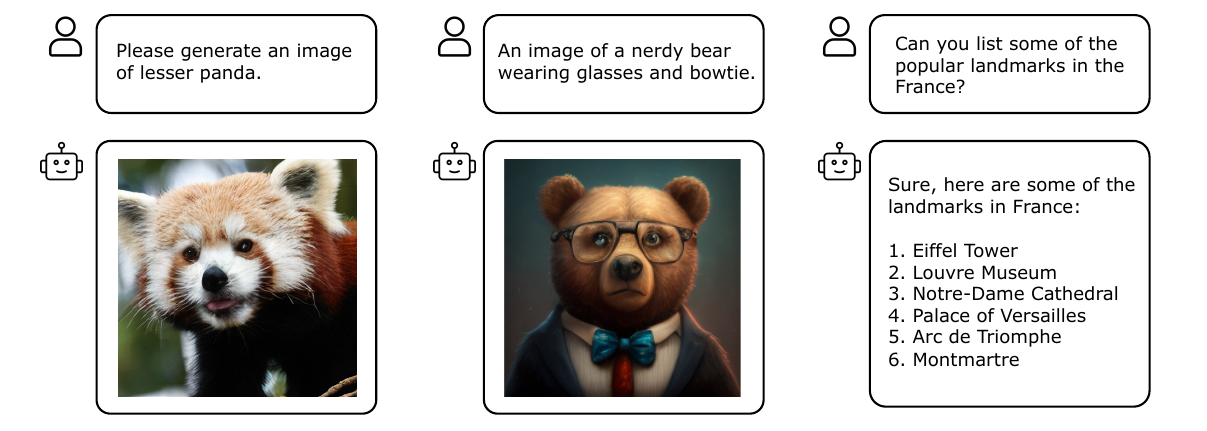}
    \caption{Examples of \method~for both image generation and text generation tasks. Left: Class-conditional image generation. Middle: Text-to-image generation. Right: Text-to-text generation.}
    \label{fig:demo}
\end{figure}

We train our model on two tasks illustrated
in Figure~\ref{fig:demo}:
autoregression for text-to-text generation and diffusion for text-to-image generation.
We use a pretrained LLM as the transformer backbone to gain language understanding capabilities.
Experiments demonstrate that our method achieves comparable image generation performance to current state-of-the-art methods
as well as maintains the text generation capability.

\section{Related Works}
\noindent\textbf{Diffusion models.}
Diffusion models~\citep{sohl2015deep,ho2020denoising,rombach2022high,song2020denoising} have demonstrated outstanding performance in generating high-quality images.
These models show significant advantages in terms of stability and scalability.
Diffusion models~\citep{nichol2021glide,saharia2022photorealistic} 
for text-to-image generation
incorporate pretrained text encoders,
such as CLIP~\citep{radford2021learning}, Flan-T5~\citep{chung2024scaling}, or LLaMA~\citep{touvron2023llama},
or  
use LLMs to encode the text~\citep{koh2024generating} 
as the condition.

Recently, the architecture of diffusion models has been shifting from U-Net architectures to transformer-based architectures~\citep{peebles2023scalable,podell2023sdxl,gao2024lumina}, narrowing the gap between image generation and language understanding tasks.
This inspired us to use one transformer for both autoregression and diffusion generation.

\noindent\textbf{Autoregressive models.}
Autoregressive (AR) models are widely used in text generation tasks
and have demonstrated remarkable performance in large language models (LLMs)~\citep{brown2020language,radford2019language,touvron2023llama,team2023gemini,touvron2023llama2}.

AR models have also been applied in image generation~\citep{parmar2018image,child2019generating,ramesh2021zero,chang2023muse,ding2022cogview2,saharia2022photorealistic,li2024autoregressive}.
The AR models usually use visual tokenizers, such as VQ-VAE~\citep{van2017neural} or VQ-GAN~\citep{esser2021taming}, to discretize
continuous image features into discrete tokens.
Though AR models perform well for image generation~\citep{ramesh2022hierarchical,yu2022scaling,chen2020generative},
the overall quality is still inferior to diffusion models.

\begin{figure}[t]
  \centering
  \footnotesize
\includegraphics[scale=0.58]{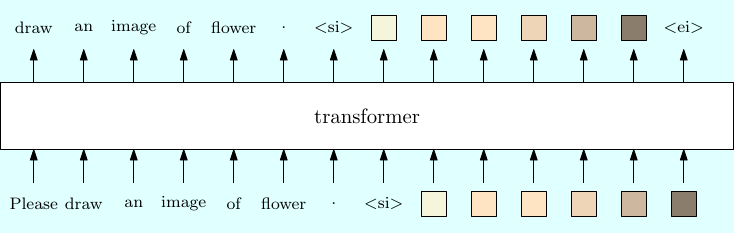}~~
\includegraphics[scale=0.58]{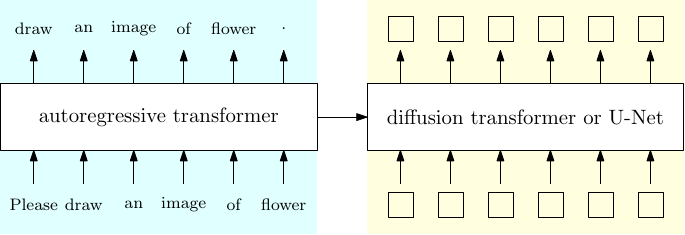}

   \caption{Left:
   A single autoregressive transformer
   for both text generation
   and visual generation.
   Example methods inlcude {Chameleon~\citep{team2024chameleon} and LlamaGen~\citep{sun2024autoregressive}}.
   Right: 
   {One transformer is for autoregressive text generation, and the output embeddings are sent to another model
   for diffusion-based text-to-image generation~\citep{ge2024seed,wu2023next}.}
   }
   \label{fig:onecol}
\end{figure}

\noindent\textbf{Unified models.}
There has been a trend in the community towards a unified model for text-to-text generation and text-to-image generation.
Recent works~\citep{team2024chameleon,he2024marsmixtureautoregressivemodels} 
propose to use a single autoregressive model
for text-to-text generation
and text-to-image generation,
illustrated in Figure~\ref{fig:onecol} Left.
These methods still use autoregression for image generation
and thus still under-perform diffusion models.
Prior works simply adopt
an additional network for diffusion
to generate high-quality images~\cite{ge2024seed,wu2023next},
which is illustrated in
Figure~\ref{fig:onecol} Right.

\section{Method}

\subsection{Preliminaries}

\noindent\textbf{Autoregressive transformer.}
An autoregressive model factorizes the joint distribution autoregressively
over a sequence of high-dimensional data $\{x_1, x_2, \dots, x_n\}$:
\begin{align}
    p({x}_1, {x}_2, 
    \dots, {x}_n)
    = \prod\nolimits_{i=1}^n p({x}_i | {x}_{<i}; \vtheta).
\end{align}
The model, which is parameterized by $\vtheta$, is trained by
minimizing the negative log-likelihood:
\begin{align}
    \mathbb{E}_{\mathbf{x} \in {\mathcal{X}}}
    [-\log p(\mathbf{x})]
    = \mathbb{E}_{\mathbf{x} \in {\mathcal{X}}}
    [-\sum\nolimits_{i=1}^n\log p(x_i | x_{<i}; \vtheta)].
\end{align}
For inference,
the model predicts the data one by one:
predict $x_1$, $\dots$, $x_i$, $\dots$ until $x_n$ by sampling from $p(x_1; \vtheta)$, $\dots$, $p(x_i | x_{<i}; \vtheta)$, $\dots$, $p(x_n | x_{<n}; \vtheta)$.

The transformer decoder is used in models like GPT and LLaMA,
for the implementation. 
The decoder takes token embeddings as input and produces an embedding for each position. 
The autoregressive transformer is similar to a standard transformer decoder, with the primary difference being the use of a causal attention mask. 
This mask ensures that each position only attends to previous positions, maintaining the autoregressive property.

\noindent\textbf{Diffusion transformer.}
Diffusion models
consist of a 
diffusion process, 
which gradually adds Gaussian noise to the data $x_0$:
\begin{align}
   x_t = \sqrt{\bar{\alpha}_t}
   x_0 + \sqrt{1 - \bar{\alpha}_t} \epsilon,
\end{align}
where $\bar{\alpha}_t$
are parameters derived
from the variance schedule,
and $\epsilon$ is the Gaussian noise. 
Diffusion models are trained
to learn the denoising process:
\begin{align}
    p_{\vtheta}(x_{t-1} | x_{t})
    = \mathcal{N}(x_{t-1};
    \mu_{\vtheta}(x_t),
    \sigma_{\vtheta}(x_t)).
\end{align}
The DDPM~\citep{ho2020denoising} transfers the problem
through reparameterization 
to learn a noise prediction network
$\epsilon_{\vtheta}$
by minimizing the difference
between the predicted noise $\epsilon_{\vtheta}({x}_t)$
and the ground-truth noise $\epsilon$:
\begin{align}
    \mathcal{L}_{\texttt{simple}}
    = \|\epsilon_{\vtheta}({x}_t) - \epsilon\|_2^2.
\end{align}

U-Net~\citep{ronneberger2015unetconvolutionalnetworksbiomedical} is widely used 
as the backbone for 
the noise prediction network $\epsilon_{\vtheta}$.
Latent diffusion models~\citep{rombach2022high}
train the model
in the latent representation space
that is formed
by learning a variational autoencoder (VAE) ~\citep{kingma2013auto}
to compress images.

Recently, Diffusion Transformers (DiTs)~\citep{peebles2023scalable}, which our approach builds upon, train latent diffusion models using a transformer.
We adopt the variant
based on the DiT block with in-context conditioning.
The architecture utilizes standard self-attention over condition embeddings (for text and timesteps) and latent embeddings.
The self-attention is bidirectional,
meaning that there is no mask
or equivalently the mask is an all-ones matrix.

\setlength\dashlinedash{0.2pt}
\setlength\dashlinegap{1.5pt}
\setlength\arrayrulewidth{0.3pt}

\subsection{MonoFormer}
Our approach is built upon the common-used 
large language model architecture,
and uses one transformer
for both autoregression-based text-to-text generation and 
diffusion-based text-to-image generation.

For text-to-text generation, the input text is processed as a sequence of text token embeddings, which are fed into the transformer.
The transformer autoregressively generates output embeddings at each position, which is then parsed into text tokens.

For text-to-image generation,
the noised latents
are sent to the transformer
to generate embeddings
at all positions simultaneously.
This process is iterated over multiple timesteps
in a cascade manner,
following the standard denoising
diffusion pipeline.
The final output embeddings 
are decoded to an image through a VAE decoder.

\noindent\textbf{Training.}
\label{sec:training}
The transformer is trained
using the autoregression loss
for text-to-text generation,
and the diffusion loss
for text-to-image generation.
We adopt the standard LLM transformer architecture.
The transformer is 
composed of transformer blocks.
Each block is formed
by the FFN and the masked attention:
\begin{align}
    \operatorname{masked-attention}(\mathbf{Q},
    \mathbf{K},
    \mathbf{V})
    = \operatorname{softmax}(\frac{\mathbf{Q}
    \mathbf{K}^{\top}}{\sqrt{d}}
    \odot \mathbf{M})\mathbf{V},
\end{align}
where $\mathbf{M}$
is the attention mask,
$\odot$ is the element-wise product,
and $d$ is the dimension.
The key difference
between autoregression
and diffusion
lies in the mask used.

We adopt the standard
causal attention mask
(upper triangular mask)
for optimizing the autoregression loss:
\begin{align}
  \mathbf{M}_{\mathtt{AR}} =  \begin{bmatrix}
1 & -\infty & -\infty & -\infty & -\infty & -\infty & -\infty \\
1 & 1 & -\infty & -\infty & -\infty & -\infty & -\infty \\
1 & 1 & 1 & -\infty & -\infty & -\infty & -\infty \\
1 & 1 & 1 & 1 & -\infty & -\infty & -\infty \\
1 & 1 & 1 & 1 & 1 & -\infty & -\infty \\
1 & 1 & 1 & 1 & 1 & 1 & -\infty \\
1 & 1 & 1 & 1 & 1 & 1 & 1
\end{bmatrix}.
\end{align}

The transformer takes the text embedding $\mathbf{t}_o$ as input
and applies the causal attention mask $\mathbf{M}_{\mathtt{AR}}$ to mask past tokens. 
The output embeddings are then used to predict the tokens through the autoregression head $\mathcal{H}_{\mathtt{AR}}$:
\begin{align}
\bar{\mathbf{t}}_o = \mathcal{H}_{\mathtt{AR}}(\operatorname{transformer}(\mathbf{t}_o, \mathbf{M}_{\mathtt{AR}}; \vtheta)).
\end{align}

Optimizing the diffusion loss 
is similar.
The major difference 
lies in the attention mask.
Unlike in text generation, where text tokens see only past tokens,
image tokens see both past text tokens and future image tokens.
A bidirectional attention mask is adopted. 
We illustrate the attention mask
using an example
with three text tokens and four image tokens:
\begin{align}
 \mathbf{M}_{\mathtt{Di}} = \begin{bmatrix}
1 & -\infty & -\infty & -\infty & -\infty & -\infty & -\infty \\
1 & 1 & -\infty & -\infty & -\infty & -\infty & -\infty \\
1 & 1 & 1 & -\infty & -\infty & -\infty & -\infty \\
\hdashline
1 & 1 & 1 & 1 & 1 & 1 & 1\\
1 & 1 & 1 & 1 & 1 & 1 & 1\\
1 & 1 & 1 & 1 & 1 & 1 & 1\\
1 & 1 & 1 & 1 & 1 & 1 & 1\\
\end{bmatrix}.
\end{align}

The transformer takes the embeddings
of the text $\mathbf{t}_{o}$
and noised latent embeddings
$\mathbf{z}_t$
as input,
processing them with the attention mask $\mathbf{M}_{\mathtt{Di}}$.
It outputs embeddings that are then passed through the diffusion head $\mathcal{H}_{\mathtt{Di}}$ to predict the noise:
\begin{align}
\bar{\epsilon} = \mathcal{H}_{\mathtt{Di}}
(\operatorname{transformer}(\mathbf{t}_o, \mathbf{z}_t, \mathbf{M}_{\mathtt{Di}}; \vtheta)).
\end{align}
We use the standard diffusion loss,
as used in DiT~\citep{peebles2023scalable}, for training.

The whole loss 
is a combination
of the standard text-to-text autoregression loss 
and 
text-to-image diffusion loss:
\begin{align}
\ell_{\mathtt{AR}}(\bar{\mathbf{t}}_o, \mathbf{t}_o)
    +
    \ell_{\mathtt{Di}}(\bar{\epsilon}, \epsilon).
\end{align}

\noindent\textbf{Inference.}
For text-to-text generation,
the inference is a standard autoregression process,
using the trained {$\operatorname{transformer}$}
to predict next tokens
one by one.

For text-to-image generation,
we follow DiT~\citep{peebles2023scalable}.
Once the image start token \texttt{<si>} is generated in the autoregression process, the diffusion process begins.
We input the Gaussian noise
as the initial noised latent,
to the $\operatorname{transformer}$,
predicting the noise
that is used to reduce
the noise.
The noise reduction process
is iterated for multiple timesteps
to generate the image.
The denoising process
is almost the same
as that of
the {in-context version of DiT}~\citep{peebles2023scalable}
with a slight difference:
Our approach uses a
combination of causal attention mask
(for text tokens)
and the bidirectional attention mask
(for image tokens).
In contrast, the in-context version of DiT uses a bidirectional attention mask for both text and image tokens.

\noindent\textbf{Architecture for diffusion.}
We perform diffusion in the latent space.
We use the VAE encoder 
to map each image patch
to a continuous representation
followed by a linear projection to generate
latent representations
for latent diffusion.

The noised latent embeddings are added with positional embeddings, using the sine-cosine version
Following DiT~\citep{peebles2023scalable}, we embed the input timestep using a $256$-dimensional frequency embedding, followed by a two-layer MLP with SiLU activations.
The time embeddings are combined
with the noised latent embeddings through an AdaLN layer.
The combined embeddings
are fed into the transformer
for predicting the noise,
followed by denoising the noised embeddings.

The diffusion head $\mathcal{H}_{\mathtt{Di}}$ is implemented following DiT, which consists of a layer norm followed by a linear layer and SiLU activation.

\section{Experiments}
\subsection{Setup}

\textbf{Implementation details.} 
We leverage the pretrained variational autoencoder (VAE) from Stable Diffusion~\citep{rombach2022high}
to encode the image.
The VAE encoder downsamples
the image by 
a factor of $8$, 
generating a latent representation of dimension $4$.
Following DiT \citep{peebles2023scalable}, we patchify the latent representation using patch size of $2 \times 2$.
The latent representation goes through
a linear projection layer
for matching the dimension
of the representation that is fed
into the transformer.
The linear projection layer for aligning the latent representation dimensions is initialized to zero.
The linear layers in the time embedding projector
are initialized to zero.
The AdaLN parameters
for combining time embeddings are initialized using a normal distribution.
{
The linear layers in the diffusion head $\mathcal{H}_\mathtt{Di}$ are initialized using a normal distribution too.}
We initialize the transformer
using TinyLlama-1.1B v1.0~\citep{zhang2024tinyllama}, which is pre-trained on 3T tokens and fine-tuned on UltraChat~\citep{ding2023enhancing} dataset.

We use the AdamW optimizer without weight decay, and the learning rate is set to a constant value of 1e-4.
We maintain an exponential moving average of the \method~weights throughout the training process with a decay rate of $0.9999$.
We retain the diffusion hyper-parameters from DiT~\citep{peebles2023scalable}, 
using $1000$ timesteps linear variance schedule ranging from $1\times 10^{-4}$ to  $2\times 10^{-2}$, and parameterization of the covariance $\Sigma_\theta$.

We train the model on the ImageNet~\citep{deng2009imagenet} dataset for class-conditional generation.
The category names are converted into text to form the text prompt in ImageNet, such as ``Please generate an image of \texttt{[category]}", ``An image of \texttt{[category]}", etc.
We train the model on the JourneyDB~\citep{sun2024journeydb} and UltraChat~\citep{ding2023enhancing} for text-to-image generation and text generation.
Considering that the diffusion task is more difficult,
the ratio
of the numbers of image generation samples and text generation samples is $9:1$.
The Global batch size is $1024$.

\noindent\textbf{Classifier-free guidance.}
We adopt classifier-free guidance~\citep{ho2022classifier} for text-to-image generation.
In the training stage, we randomly drop the text tokens for unconditional generation.
We set the drop probability to $1/10$.
The final latent output is computed by combining the unconditional and conditional outputs and using a guidance factor to control the scale of the guidance.

\subsection{Experiment Results}

\textbf{Image generation.} 
We evaluate the text-to-image generation performance on the ImageNet~\citep{deng2009imagenet} dataset under resolution of $256\times 256$.
The evaluation metrics 
include:
the Fréchet Inception Distance (FID),
the Inception Score (IS), 
and Precision/Recall.
In inference, we use a classifier-free guidance scale of 1.5.
We report the comparison
to representative diffusion models and unified autoregressive models. 
The comparison results are shown 
in Table~\ref{tab:comparison_imagenet}.
{Our method achieves comparable results with recent diffusion-based or AR-based methods. 
It outperforms the AR-based LlamaGen-3B~\citep{sun2024autoregressive} in FiD with fewer parameters, while being only 0.3 lower than the state-of-the-art diffusion-based method DiT-XL/2~\citep{peebles2023scalable} in FID.}

\begin{table}[t]
    \centering
    \footnotesize
    \setlength{\tabcolsep}{6.5pt}
    \renewcommand{\arraystretch}{1.25}
    \caption{\textbf{Performance on ImageNet 256$\times$256 benchmark.}}
    \begin{tabular}{l | c  c | c  c  c  c  c}
        \shline
        \textbf{Method} & \textbf{Arch} & \textbf{\#Params} & \textbf{FID$\downarrow$} & \textbf{IS$\uparrow$} & \textbf{Precision$\uparrow$} & \textbf{Recall$\uparrow$} \\
        \hline
        ADM~\citep{dhariwal2021diffusion}       &Diff&554M&10.94&101.0&0.69&0.63\\
        CDM~\citep{ho2022cascaded}       & Diff&-&4.88& 158.7&           -          &            -      \\
        LDM~\citep{rombach2022high}       & Diff    &    400M      &       3.60&          147.6    &            0.87         &        0.68          \\
        DiT-XL/2~\citep{peebles2023scalable}  & Diff    &     675M     &         2.27        &       278.2      &          0.83           &          0.57        \\
        \hline
        VQGAN~\citep{esser2021taming}     & AR           &      227M    &         18.65        &        80.4      &        0.78             &        0.26          \\
        ViT-VQGAN~\citep{yu2021vector} & AR           &      1.7B    &          4.17      &    175.1          &          -           &           -       \\
        LlamaGen-B~\citep{sun2024autoregressive}  & AR     &   111M    &      5.46  &         193.6         &      0.83        &            0.45                         \\
        LlamaGen-3B~\citep{sun2024autoregressive}  & AR     &   3.1B    &      2.81  &         311.5         &      0.84        &            0.54                         \\
        \hline
        \method & AR+Diff &      1.1B    &      2.57  &     272.6     &        0.84            &  0.56  \\
        \shline
    \end{tabular}
    \label{tab:comparison_imagenet}
\end{table}

\begin{table}[b]
\centering
\footnotesize
\setlength{\tabcolsep}{2pt}
\renewcommand{\arraystretch}{1.25}
\caption{\textbf{Performance on commonsense reasoning tasks.}}
\begin{tabular}{l|cccccccc}
\shline
\textbf{Model} & \textbf{HellaSwag} & \textbf{OBQA} & \textbf{WinoGrande} & \textbf{ARC-C} & \textbf{ARC-E} & \textbf{BoolQ} & \textbf{PIQA} & \textbf{avg} \\ \hline
{Pythia~\citep{biderman2023pythia}} & 52.01 & 33.20 & 57.38 & 28.50 & 54.00 & 63.27 & 70.95 & 51.33 \\ 
{TinyLlama~\citep{zhang2024tinyllama}} & 59.20 & 36.00 & 59.12 & 30.12 & 55.25 & 57.83 & 73.29 & 52.97 \\ \hline
{\method} & 50.62 & 37.20 & 56.91 & 31.48 & 48.19 & 62.29 & 71.16 & 51.12 \\ \shline
\end{tabular}
\label{tab:text2text_comparison}
\end{table}

\textbf{Text generation.}
\label{sec:t2t}
We evaluate \method~on a diverse set of commonsense reasoning tasks and compare it with the baseline model, TinyLlama~\citep{zhang2024tinyllama}.
We evaluate our method on the following tasks: HellaSwag~\citep{zellers2019hellaswag}, OpenBookQA~\citep{mihaylov2018can}, WinoGrande~\citep{sakaguchi2021winogrande}, ARC-Easy and ARC-Challenge~\citep{clark2018think}, BoolQ~\citep{clark2019boolq}, and PIQA~\citep{bisk2020piqa}.

Table~\ref{tab:text2text_comparison}
shows that \method~achieves performance comparable to the TinyLlama baseline, with slight drops on certain benchmarks (average score deceases from $52.97$ to $51.12$).
{
The slight performance drops may be attributed to the mixed training with the image generation dataset.
We believe the performance can be further improved when more language data are incorporated in training, we leave this for future exploration.}

\begin{figure}[t]
    \centering
    \includegraphics[height=3.8cm]{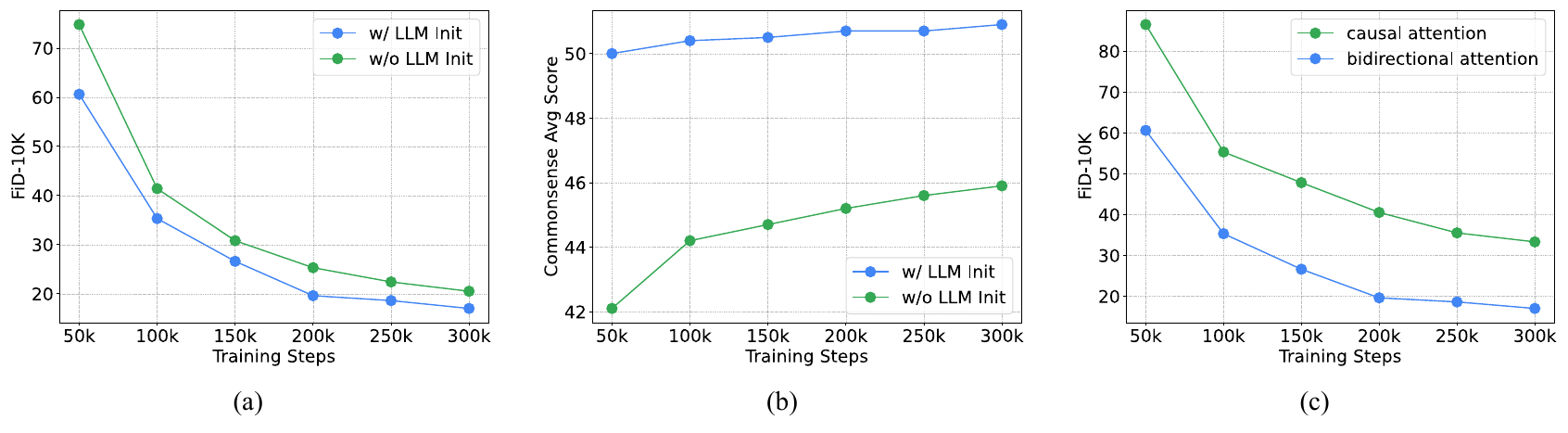}
    \vspace{-5mm}
    \caption{{{(a) The effect of transformer initialization for image generation, measured using the FiD-10K metric on ImageNet. (b) The effect of transformer initialization for text generation, measured by the average commonsense reasoning score. (c) The effect of bidirectional attention mask for image generation.}}}
    \label{fig:abl}
    \vspace{5mm}
\end{figure}

\subsection{Ablation Study}
\textbf{Transformer initialization.}
We conduct an ablation study to study the impact of using pretrained LLMs for transformer initialization.
We compare the performance of \method~with and without pretrained LLM initialization on two tasks: image generation on the ImageNet 256$\times$256 benchmark and language understanding on commonsense reasoning benchmarks.
Performance is evaluated using the FiD score for ImageNet and the average score across 6 commonsense reasoning tasks, as described in Section~\ref{sec:t2t}.
As shown in Figure~\ref{fig:abl} (a), \method~with pretrained LLM initialization significantly outperforms the counterpart in commonsense reasoning.
As shown in Figure~\ref{fig:abl} (b), on the image generation benchmark, \method~with LLM initialization also shows superior performance. 
We attribute this improvement to the pretrained LLM's ability to better understand prompts, thereby benefiting text-to-image generation tasks.

\textbf{Bidirectional attention for diffusion.}
We study the effect of applying bidirectional attention masks among noised latent tokens, which are used for diffusion-based generation.
We respectively use causal attention mask and bidirectional attention mask for the noised latent tokens and compare their performance on the ImageNet 256$\times$256 benchmark.
As shown in Figure~\ref{fig:abl} (c),
the performance of \method~with bidirectional attention outperforms \method~with causal attention mask, demonstrating the importance of bidirectional attention mask for image generation.

\section{Conclusion}
This paper shows that discrete autoregression 
and continuous diffusion are able to share one transformer.
Experiments validate that
sharing the transformer 
is able to achieve good performance
for text-to-text generation
and text-to-image generation
that is comparable to 
not sharing the transformer.

\bibliography{iclr2025_conference}
\bibliographystyle{iclr2025_conference}

\newpage
\clearpage

\appendix

\section{Image Generation Results}
\begin{figure}
    \centering
    \includegraphics[width=1.0\linewidth]{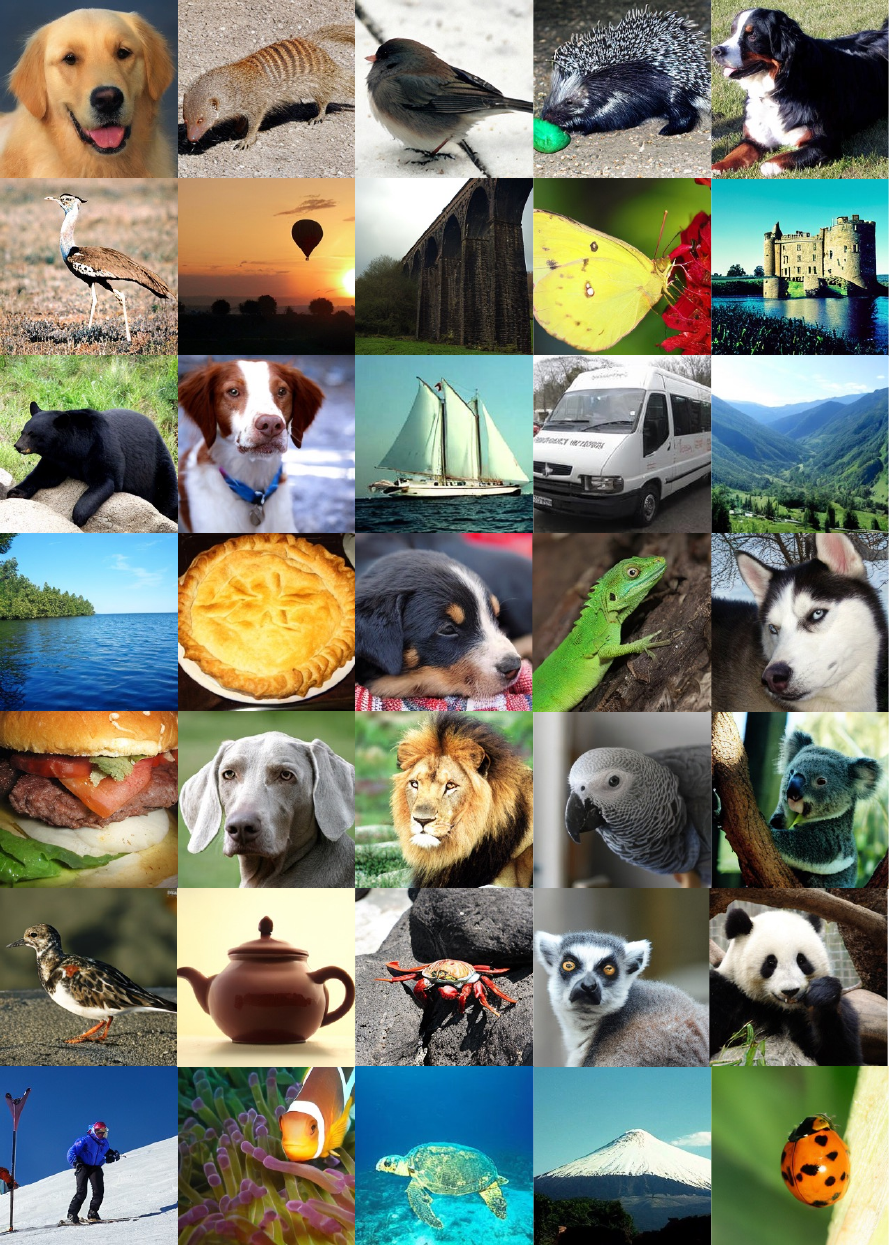}
    \caption{\textbf{Example results of class-conditional image generation on ImageNet.} We use 250 sampling steps and a classifier-free guidance scale of 4.0.}
    \label{fig:vis_imagenet}
\end{figure}

\begin{figure}
    \centering
    \includegraphics[width=1.0\linewidth]{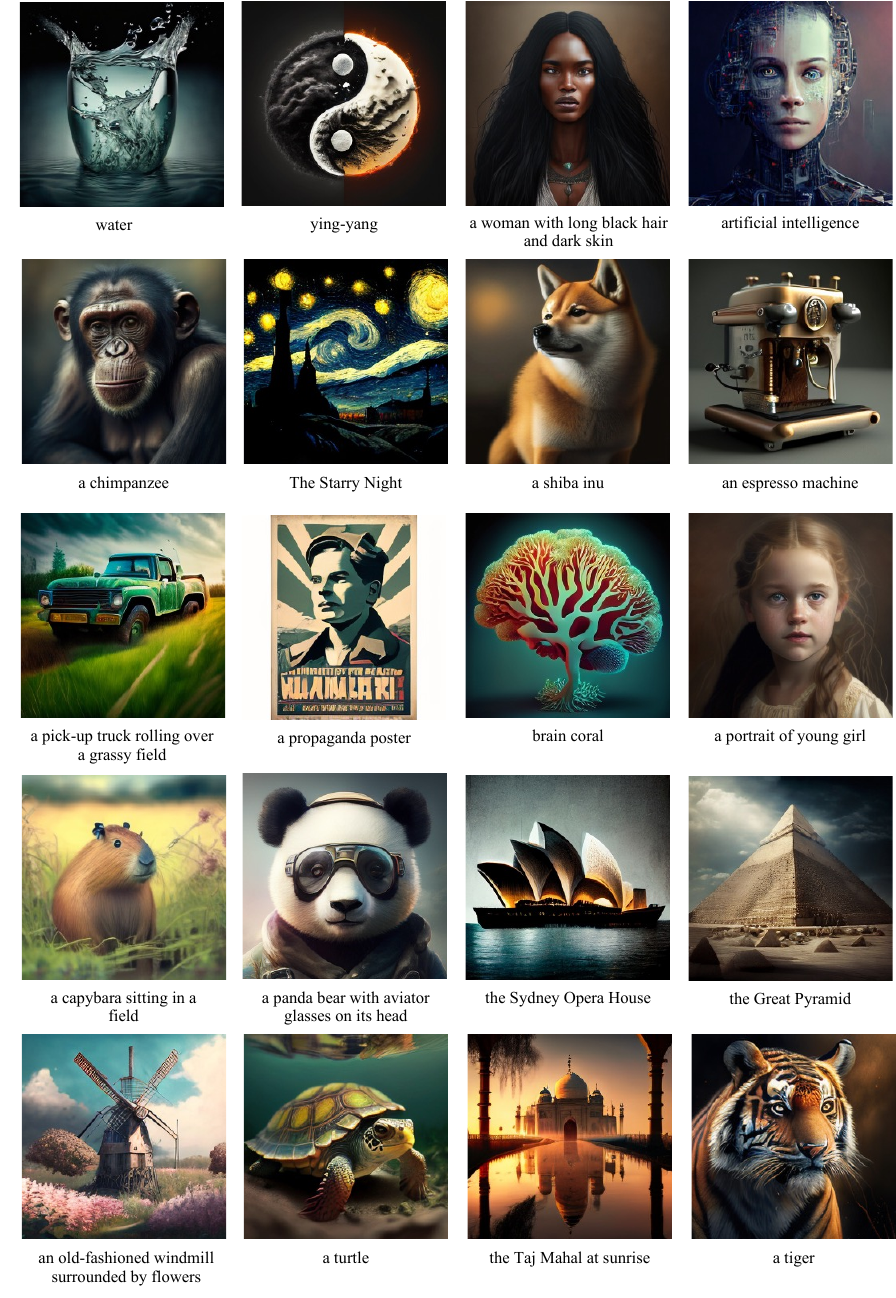}
    \caption{\textbf{Example results of text-to-image generation.} The prompts are from Parti prompts~\citep{yu2022scaling}. We use 250
sampling steps and a classifier-free guidance scale of 4.0.}
    \label{fig:vis_t2i}
\end{figure}

We present class-conditional image generation results in Figure~\ref{fig:vis_imagenet}.
We use a classifier-free guidance scale of 4 and sample 250 steps using DPM-Solver.
We further train our model on the JourneyDB~\citep{sun2024journeydb} dataset and the UltraChat~\citep{ding2023enhancing} dataset for free text-to-image generation.
Figure~\ref{fig:vis_t2i} showcases the results generated using Parti prompts~\citep{yu2022scaling}, with a classifier-free guidance scale of 4 and 250 sampling steps.

\section{Extension to Multi-modality Understanding}
Our approach can be easily extended
for multi-modality understanding,
for example, vision-language understanding.
In the case that the data sample is an interleaved image-text sequence,
there are two choices for extracting the image representation for understanding.
One choice is that the image is only processed by diffusion,
and we choose to use the transformer with a timestep near zero for image representation extraction.
The other choice is that
the image is processed 
in an autoregressive manner for understanding
and diffusion is only for generation.

\section{Diccussion with concurrent work}
{
There are concurrent works 
with similar ideas,
such as Transfusion~\citep{zhou2024transfusionpredicttokendiffuse} and Show-o~\citep{xie2024showosingletransformerunify}.
Show-o employs discrete diffusion. 
Differently, 
our approach uses continuous diffusion model.
Transfusion is more similar to our method,
and trained from scratch.
Our experiments demonstrate that initializing transformer with a large language model is helpful
for training.}

\end{document}

%% file: math_commands.tex
\usepackage{amsmath,amsfonts,bm}

\def\eqref#1{equation~\ref{#1}}

\def\1{\bm{1}}

\def\vtheta{{\bm{\theta}}}

\DeclareMathAlphabet{\mathsfit}{\encodingdefault}{\sfdefault}{m}{sl}
\SetMathAlphabet{\mathsfit}{bold}{\encodingdefault}{\sfdefault}{bx}{n}